\documentclass[lettersize,journal]{IEEEtran}

\usepackage{amsmath,amssymb,amsfonts}
\usepackage{algorithm}
\usepackage{algorithmic}
\usepackage{array}
\usepackage[caption=false,font=normalsize,labelfont=sf,textfont=sf]{subfig}
\usepackage{textcomp}
\usepackage{stfloats}
\usepackage{url}
\usepackage{verbatim}
\usepackage{graphicx}
\graphicspath{{original/}}
\usepackage{cite}
\usepackage{booktabs}
\usepackage{multirow}
\usepackage{xspace}
\usepackage[hidelinks]{hyperref}

\newcommand{\eg}{\emph{e.g.}\xspace}
\newcommand{\ie}{\emph{i.e.}\xspace}

\newcommand{\ACC}{\text{ACC}}
\newcommand{\MFone}{\text{MF1}}
\newcommand{\AAA}{\text{AAA}}
\newcommand{\AAFone}{\text{AAF1}}

\title{Adaptive Data Compression and Reconstruction for Memory-Bounded EEG Continual Learning}

\author{Chengcheng~Xie%
\thanks{Chengcheng Xie is with Beihang University (e-mail: \texttt{sonicrainboomxcc@buaa.edu.cn}).}}

\begin{document}
\maketitle

\begin{abstract}
Electroencephalography (EEG) signals provide millisecond-level temporal resolution but their analysis is limited by remarkable noise and inter-subject variability, making robust personalization difficult under limited annotations. Unsupervised Individual Continual Learning (UICL) has been proposed to address this practical challenge, where a model pretrained on a labeled cohort must adapt online to unlabeled subject streams under strict memory constraints. However, existing UICL methods typically store full past samples, which undermine the continual learning goal of avoiding retraining. Observing that EEG signals exhibit well-structured morphologies to be exploited via morphology-aware selection, compression, and reconstruction, here we propose \textbf{A}daptive \textbf{Da}ta \textbf{Co}mpression and \textbf{Re}construction (ADaCoRe) for UICL. This is a memory-efficient pipeline composed of saliency-driven keyframe protection, rational polyphase compression, adjoint reconstruction with verbatim overwrite on protected indices, and prototype-confidence selection for adaptive exemplar maintenance. Across three representative benchmarks, ADaCoRe consistently outperforms recent strong baselines under tight buffer regimes (\eg, the performance gains are at least +2.7 and +15.3 ACC on ISRUC and FACED datasets, respectively).
Ablation studies quantify compression-fidelity trade-offs and highlight the contribution of each design, while visualizations confirm the preservation of key EEG morphology during compression and reconstruction.
\end{abstract}



\section{Introduction}
\IEEEPARstart{E}{lectroencephalography} (EEG) is a noninvasive, cost-effective, and high-temporal-resolution modality widely used in real-world applications such as at-home monitoring~\cite{Milne-Ives2023_at_home_EEG_monitoring,Weisdorf2019_ultralong_subcutaneous_EEG}, brain-computer interfaces (BCIs)~\cite{McFarland2017_EEGbased_BCI}, and sleep staging~\cite{mathias2019utime}. These applications increasingly rely on subject-personalized models to handle substantial inter-individual variability. However, labeled EEG data are often scarce or delayed, especially in long-term or in-the-wild deployments. These considerations motivate the need for models that can adapt continually to new subjects and recording conditions without access to labels or unbounded storage.

\begin{figure}[t]
  \centering
  \includegraphics[width=0.85\linewidth]{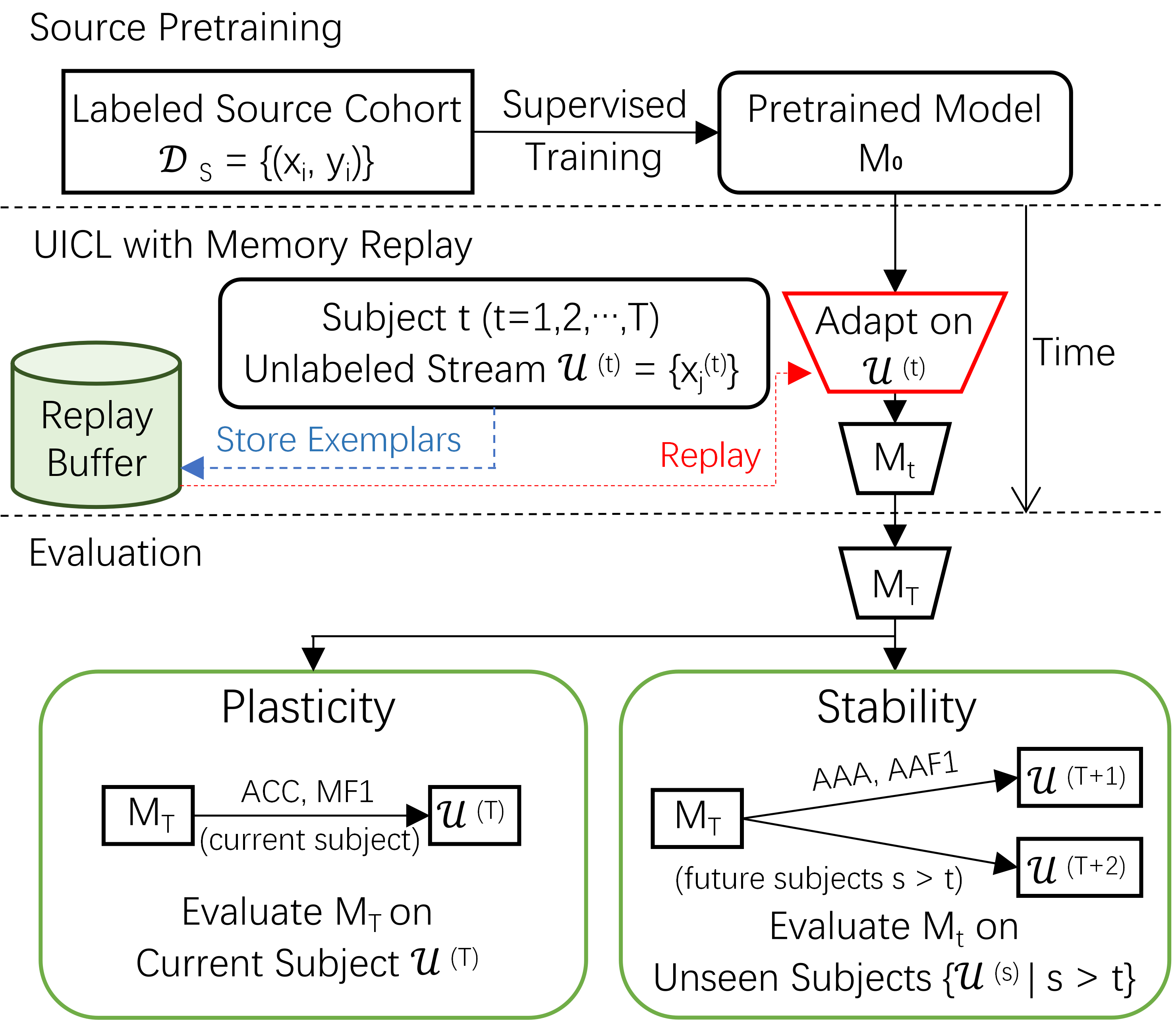}
  \caption{UICL. A source-pretrained model $M_0$ adapts to a sequence of unlabeled subject streams $\{\mathcal{U}^{(t)}\}_{t=1}^{T}$ using a memory-bounded replay buffer, producing $M_t$ after subject $t$. Plasticity and stability are measured on the current and future subjects, respectively.}
  \label{fig:uicl_pipeline}
\end{figure}

This problem, referred to as \textit{Unsupervised Individual Continual Learning} (UICL, Fig.~\ref{fig:uicl_pipeline}), presents several unique challenges: data arrive as subject-specific streams, supervision is unavailable, and memory is tightly constrained. Existing methods often rely on replay-based continual learning, storing previously seen examples in a memory buffer for future rehearsal~\cite{rebuffi2017icarl,lopezpaz2017gem}. However, these methods typically assume generous or unbounded buffers and store full-resolution EEG segments indiscriminately. This incurs inefficient use of memory, redundancy from storing non-informative signals, and potential overfitting to noisy or low-saliency segments. Moreover, they seldom exploit domain-specific signal characteristics such as rhythmicity, spectral structure, or morphological saliency, despite their central role in EEG interpretation~\cite{aasm2007manual}.

In contrast, our work builds on the insight that EEG signals exhibit well-defined morphological patterns (\eg, sleep spindles, K-complexes, and event-related potentials), which can be selectively preserved and reconstructed with high fidelity. We propose \textbf{ADaCoRe}, a signal-aware, memory-efficient pipeline for UICL in EEG applications. ADaCoRe comprises four coordinated components: (1) \emph{saliency-driven keyframe protection} to retain critical segments at full resolution; (2) \emph{rational polyphase compression} with anti-aliasing for efficient downsampling of non-keyframes; (3) \emph{adjoint reconstruction} that overwrites protected indices verbatim during replay; and (4) \emph{prototype-confidence selection} that maintains a compact yet informative buffer using task-relevant confidence and diversity heuristics. Together, these modules enable adaptive compression that respects EEG morphology, prevents aliasing, and prioritizes exemplars that are both representative and reliable.

We evaluate ADaCoRe on three representative benchmarks under strict memory constraints and observe consistent improvements over strong baselines, including the state-of-the-art method BrainUICL~\cite{zhou2025brainuicl}. ADaCoRe yields significantly better plasticity and stability, maintaining high performance even with significantly reduced buffer sizes. For example, the performance gains are at least +2.7 and +15.3 ACC on ISRUC and FACED datasets, respectively. Detailed ablations examine the effects of compression ratio, saliency coverage, protection radius, and selection strategies. Visual analyses of reconstructed signals confirm that ADaCoRe successfully preserves essential EEG morphologies during replay.

Our contributions are summarized as follows: (1) A morphology-aware compression and reconstruction pipeline that preserves salient EEG features under strict memory constraints. (2) A prototype-confidence selection strategy that maintains diverse and reliable exemplars for memory-efficient replay. (3) Significant performance gains over recent strong baselines for UICL in EEG applications.

\begin{figure*}[t]
  \centering
  \includegraphics[width=0.85\textwidth]{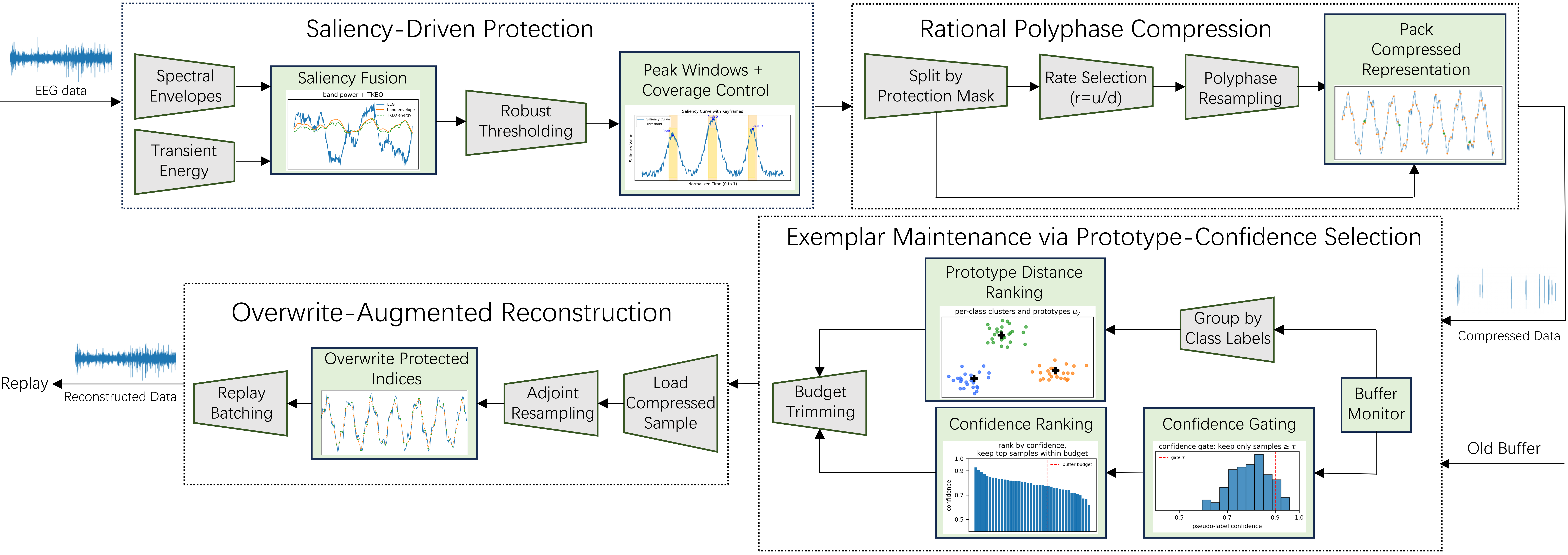}
  \caption{\textbf{Overview of ADaCoRe.}
  Saliency detection (\S\ref{subsec:saliency}) selects protected indices $\mathcal{P}$. Non-protected regions are compressed via rational polyphase resampling (\S\ref{subsec:compression}). Replay uses adjoint reconstruction with verbatim overwrite of protected samples (\S\ref{subsec:reconstruction}). Exemplar maintenance respects class prototypes and pseudo-label confidence (\S\ref{subsec:pcs}).}
  \label{fig:adacore_overview}
\end{figure*}

\section{Related Work}
\label{sec:related-work}

\textbf{Continual Learning (CL).}
CL aims to learn from nonstationary data streams without full access to previous training samples. Representative methods include parameter regularization~\cite{kirkpatrick2017ewc,zenke2017si}, knowledge distillation~\cite{li2017lwf}, and memory replay~\cite{rebuffi2017icarl,lopezpaz2017gem}. However, most CL research targets supervised vision with ample labels and large buffers, limiting applicability to label-scarce, memory-constrained EEG scenarios. UICL has been proposed specifically for EEG adaptation to unlabeled, subject-specific streams with bounded memory. BrainUICL~\cite{zhou2025brainuicl} is a recent representative baseline that we compare against.

\textbf{Continual Domain Adaptation (CDA) and Test-Time Adaptation (TTA).}
CDA focuses on adapting a source-pretrained model to a sequence of unlabeled target domains. Representative strategies include discrepancy alignment and adversarial domain adaptation~\cite{tzeng2014ddc,ganin2016dann}. TTA methods adapt at inference time without source data, \eg, CoTTA~\cite{wang2023cotta}. Most prior CDA and TTA work is image-centric and rarely enforces realistic memory budgets or EEG-specific constraints. Our design complements these by introducing signal-aware compression and exemplar selection that can be integrated into existing adaptation frameworks.

\textbf{EEG Signal Analysis.}
EEG tasks such as sleep staging and BCI typically rely on static, fully labeled settings and AASM-compliant rules~\cite{aasm2007manual,schalk2004bci2000}. Recent works explore self-supervised time-series representation, \eg, TSTCC~\cite{eldele2021tstcc}, but continual, memory-bounded adaptation for EEG remains under-explored.

\section{Preliminaries}
\label{sec:prelim}

In this section, we introduce the problem formulation, state-of-the-art baseline, and signal processing procedures.

\subsection{Problem Formulation of UICL}
\label{subsec:prelim-problem}

We study the setting of \textit{Unsupervised Individual Continual Learning} (UICL) for EEG applications, where a model must adapt online to a sequence of previously unseen, unlabeled subjects under strict memory constraints. Concretely, the model is first pretrained on a labeled source cohort $\mathcal{D}_S = \{(x_i, y_i)\}$, where each EEG segment $x_i \in \mathbb{R}^{C \times L}$ consists of $C$ channels and $L$ time points. At deployment, the model receives a stream of unlabeled subjects $\{\mathcal{U}^{(t)}\}_{t=1}^{T}$, with each $\mathcal{U}^{(t)} = \{x_j^{(t)}\}$ denoting the samples from subject $t$. No ground-truth labels are available during adaptation.

Consistent with previous CL research, a key challenge lies in \emph{catastrophic forgetting}~\cite{mermillod2013stability}, \ie, a tendency to lose performance on earlier distributions when adapting to new ones. Specifically, UICL aims to balance \textbf{plasticity} (adaptability to the current subject) with \textbf{stability} (retention of knowledge useful for future subjects). Following BrainUICL~\cite{zhou2025brainuicl}, plasticity is evaluated on the incremental stream via \emph{Accuracy} (\ACC) and \emph{Macro-averaged F1} (\MFone), while stability is evaluated on held-out subjects via \emph{Average Anytime Accuracy} (\AAA) and \emph{Average Anytime Macro-F1} (\AAFone). Formal definitions of these metrics are provided in Appendix.

Replay-based strategies~\cite{rebuffi2017icarl,lopezpaz2017gem} help mitigate forgetting by maintaining a memory buffer $\mathcal{B}$ of past exemplars. Throughout this work we treat the buffer as a bounded store of compact representations rather than raw signals, and we measure memory usage by the number of stored timepoints (times channel count) across all buffered segments. This memory-centric view makes it straightforward to compare methods that use different compression schemes but share the same storage budget.

\subsection{State-of-the-art Baseline}
\label{subsec:prelim-brainuicl}

Our method is instantiated within the UICL pipeline introduced by the state-of-the-art BrainUICL~\cite{zhou2025brainuicl} method. In this paradigm, each subject is treated as a separate individual domain. The full set of subjects is partitioned into a labeled source set $\mathcal{D}_S$ for pretraining, an unlabeled incremental stream $\mathcal{D}_T$ for continual adaptation, and a labeled generalization set $\mathcal{D}_G$ for stability evaluation. At test time, the model observes subjects from $\mathcal{D}_T$ one by one, and after each adaptation step it is evaluated both on the current subject (plasticity) and on all subjects in $\mathcal{D}_G$ (stability), as summarized in Fig.~\ref{fig:uicl_pipeline}.

BrainUICL follows a replay-based pipeline with three main stages. First, an \emph{incremental model} $M_0$ is pretrained on $\mathcal{D}_S$ using standard cross-entropy loss. Second, for each new target subject $i$, a \emph{guiding model} $M_g$ is obtained by copying the latest incremental model $M_{i-1}$ and refined via self-supervised Contrastive Predictive Coding (CPC)~\cite{oord2018cpc} on that subject alone. The fine-tuned guiding model then produces subject-specific pseudo-labels, which are filtered by a confidence threshold to obtain a set of relatively reliable pseudo-labeled samples. Third, BrainUICL performs supervised fine-tuning of $M_{i-1}$ using a mixed mini-batch composed of current-subject samples and replayed samples from a storage center, yielding an updated model $M_i$.

The storage center in BrainUICL is implemented as a \emph{Dynamic Confident Buffer} (DCB): a pool $S_{\text{true}}$ containing a subset of labeled source-domain samples and a pool $S_{\text{pseudo}}$ containing high-confidence pseudo-labeled samples from previously seen incremental subjects. At each time step, the replay buffer $X_B$ is constructed by sampling from $S_{\text{true}}$ and $S_{\text{pseudo}}$ with a fixed ratio (\eg, 8:2) to encourage both reliable supervision and diversity. After each subject-level adaptation, the updated model $M_i$ re-predicts pseudo-labels for that subject, and samples whose prediction confidence exceeds a second threshold are added to $S_{\text{pseudo}}$.

To further stabilize learning, BrainUICL introduces a \emph{Cross-Epoch Alignment} (CEA) module. Every few epochs, the predictions of two temporal snapshots of the same incremental model (\eg, after $e$ and $e+2$ fine-tuning epochs) are aligned on replayed samples by minimizing a Kullback-Leibler divergence term between their output distributions. This consistency regularization constrains the trajectory of the model parameters, reducing overfitting to outlier individuals and improving stability on unseen subjects.

While BrainUICL effectively balances plasticity and stability under generous storage, it requires all old EEG segments at their original sampling rate and full temporal resolution. As a result, the buffer scales linearly with both the number of subjects and the segment length, which is problematic for memory-bounded on-device scenarios. Moreover, its selection strategy operates at the \emph{sample} level. Once a segment is admitted into the buffer, all its timepoints are kept. Even if only a subset carries discriminative morphology, and pseudo-label filtering alone cannot reliably remove noisy or low-saliency regions. These limitations motivate our design, which replaces raw-signal replay with \textbf{morphology-aware compression, reconstruction, and prototype-confidence selection}.

\subsection{Signal Processing Procedures}
\label{subsec:prelim-sigproc}

EEG recordings are continuous-time voltages measured at the scalp and converted to discrete-time sequences by sampling at a rate $F_s$ (\eg, 100-250\,Hz depending on the dataset). In practice, data are segmented into fixed-length epochs (\eg, 30\,s sleep epochs or shorter BCI trials), and a typical tensor has shape $[T,C,L]$ for $T$ epochs, $C$ channels, and $L$ samples per epoch. EEG preprocessing follows standard pipelines~\cite{oppenheim2010dsp}: channel selection, bandpass filtering in task-specific frequency bands, notch filtering to suppress line noise when necessary, resampling to a common rate, and per-channel normalization. Specific configurations are summarized in Appendix.

Our saliency and compression modules build on two families of time-frequency features. First, we use band-limited envelopes: each channel is filtered into canonical EEG bands (delta, theta, alpha/$\mu$, beta, sigma, etc.), and a Hilbert-transform-based analytic signal is used to obtain smooth envelopes and approximations to instantaneous band power~\cite{oppenheim2010dsp}. Second, we use transient energy features: the Teager-Kaiser energy operator emphasizes rapid amplitude changes and sharp events such as K-complexes or phasic bursts~\cite{Boudraa2018_TeagerKaiser_energy_methods}, after which we apply short-window smoothing and cross-channel aggregation. These features are combined into a scalar saliency trace over time, which drives our keyframe selection.

For memory-aware replay implementations, we perform rational resampling via polyphase filtering~\cite{oppenheim2010dsp} to compress non-protected regions and an adjoint operator to reconstruct dense sequences. The corresponding discrete-time model, convolution operators, up/down sampling definitions, and polyphase realization are detailed in Appendix, with robust median/MAD-based thresholding for peak detection~\cite{rousseeuw1993mad}.

\section{Method}
\label{sec:method}

We present \textbf{ADaCoRe}, a \emph{signal-aware replay} pipeline for memory-efficient unsupervised continual learning on EEG signals. The core challenge is to preserve salient EEG features while storing only a compact representation per subject. ADaCoRe addresses this through four coordinated components: 
(1) saliency-driven protection of morphologically important timepoints;  
(2) anti-alias compression via rational polyphase resampling;
(3) overwrite-augmented reconstruction during memory replay; and  
(4) exemplar maintenance via prototype-confidence selection.  
A contrastive predictive coding (CPC)-based teacher~\cite{oord2018cpc} provides pseudo-labels and confidence scores, and remains fixed during adaptation. The buffer stores compressed segments with lightweight metadata. 

\subsection{Saliency-Driven Protection}
\label{subsec:saliency}

Let $x \in \mathbb{R}^{N \times C}$ denote an EEG segment with $N$ timepoints and $C$ channels. We first derive a scalar saliency trace $S[n]$ designed to highlight timepoints containing strong rhythmic or transient information. This trace is computed using a lightweight unsupervised procedure that combines band-specific spectral envelopes and energy-based cues.
Specifically, we define a set of canonical EEG frequency bands $\mathcal{B}$ (delta, theta, alpha/$\mu$, beta, optionally up to 45\,Hz). For each band $b \in \mathcal{B}$, the instantaneous power $\mathrm{Pow}_b[n]$ is obtained by zero-phase bandpass filtering followed by Hilbert magnitude, then averaged across channels. Band contributions are weighted by robust statistics $w_b$ (normalized trimmed means or medians) and combined as:
\begin{equation}
S[n] = \sum_{b \in \mathcal{B}} w_b\, \mathrm{Pow}_b[n] + \gamma\, \Psi(x[n]),
\end{equation}
where $\Psi(\cdot)$ is the Teager-Kaiser operator described in Appendix and $\gamma$ is a small weighting constant~\cite{maragos1993amfm,maragos1993energy}.

We detect salient peaks where $S[n]$ exceeds the adaptive threshold
$\tau = \mathrm{Med}(S) + \kappa \cdot 1.4826\,\mathrm{MAD}(S)$,
with $\kappa \in [2,3]$ described in Appendix. Around each peak, a neighborhood of radius $\rho$ is marked as protected. To control memory use, we cap the total protected coverage to $\phi N$ samples ($\phi\!\in(0,1)$) by greedily selecting the strongest peaks until the limit is reached. The resulting set of protected indices is $\mathcal{P}$. The first and last samples are always kept for reconstruction stability.

\subsection{Rational Polyphase Compression}
\label{subsec:compression}

Outside the protected region $\mathcal{P}$, we perform rational resampling at rate $u/d$ to obtain a \emph{uniform} low-rate sequence. We apply a standard rational polyphase resampler $f_{\mathrm{RP}}$ (see Appendix for its full definition):
\[
y \;=\; f_{\mathrm{RP}}(x; u, d)\in\mathbb{R}^{\tilde N \times C},\qquad \tilde N \approx \left\lceil N\cdot \frac{u}{d}\right\rceil .
\]
We store (1) the compressed sequence $y$ (single precision), (2) the protected \emph{verbatim} samples on $\mathcal{P}$, and (3) minimal metadata $(u,d,\mathcal{P})$. Endpoints are always included in $\mathcal{P}$ to avoid extrapolation at replay.
Given a target keep ratio $r\in(0,1]$, we select $(u,d)$ via a continued-fraction rational approximation of $r$ with a bounded denominator $d\le d_{\max}$ (typically 64 or 128). Since protected coverage $|\mathcal{P}|$ slightly perturbs the realized ratio, we choose between the two closest Farey neighbors to minimize $\left||y|+|\mathcal{P}|-rN\right|$.

\subsection{Overwrite-Augmented Reconstruction}
\label{subsec:reconstruction}

For replay, we reconstruct a dense sequence by the \emph{adjoint} rate $(d,u)$ applied to the stored uniform sequence $y$:
\[
\tilde{x} \;=\; f_{\mathrm{RP}}(y;\, d, u)\in\mathbb{R}^{N\times C}.
\]
To guarantee exact fidelity over salient regions, we then overwrite protected indices by their stored verbatim values:
\[
\tilde{x}[t] \leftarrow x[t],\quad \forall\, t\in\mathcal{P}.
\]
If metadata are missing or inconsistent with $|y|$ and $N$, we fall back to per-channel linear interpolation from the joint index set $\{0,N\!-\!1\}\cup\mathcal{P}$, which always includes the endpoints to avoid extrapolation.

\begin{table*}[!t]
\caption{Overall performance. ACC and MF1 measure plasticity; AAA and AAF1 measure stability. The last three columns report buffer size as a percentage of the BrainUICL buffer for each dataset. 
}
\label{tab:overall_perf}
\centering
\resizebox{0.95\linewidth}{!}{%
\begin{tabular}{lccccccccccccccc}
\toprule
\multirow{3}{*}{Methods}
& \multicolumn{4}{c}{\textbf{ISRUC}}
& \multicolumn{4}{c}{\textbf{FACED}}
& \multicolumn{4}{c}{\textbf{PhysioNet-MI}}
& \multicolumn{3}{c}{\textbf{Buffer size (\% of BrainUICL)}} \\
\cmidrule(lr){2-5}\cmidrule(lr){6-9}\cmidrule(lr){10-13}\cmidrule(lr){14-16}
& \multicolumn{2}{c}{\textit{Plasticity}}
& \multicolumn{2}{c}{\textit{Stability}}
& \multicolumn{2}{c}{\textit{Plasticity}}
& \multicolumn{2}{c}{\textit{Stability}}
& \multicolumn{2}{c}{\textit{Plasticity}}
& \multicolumn{2}{c}{\textit{Stability}}
& ISRUC & FACED & MI \\
\cmidrule(lr){2-3}\cmidrule(lr){4-5}
\cmidrule(lr){6-7}\cmidrule(lr){8-9}
\cmidrule(lr){10-11}\cmidrule(lr){12-13}
& \ACC & \MFone & \AAA & \AAFone
& \ACC & \MFone & \AAA & \AAFone
& \ACC & \MFone & \AAA & \AAFone
& & & \\
\midrule
MMD
& 68.6 & 62.2 & 68.1 & 65.5
& 34.5 & 29.7 & 30.8 & 27.1
& 44.5 & 43.7 & 45.0 & 44.4
& -- & -- & -- \\
TSTCC
& 68.9 & 63.8 & 61.3 & 55.5
& 37.8 & 33.7 & 33.5 & 30.7
& 44.9 & 43.3 & 45.4 & 44.1
& -- & -- & -- \\
EWC
& 70.2 & 65.2 & 68.4 & 66.1
& 37.5 & 33.3 & 33.4 & 30.5
& 46.9 & 45.9 & 46.3 & 45.4
& -- & -- & -- \\
LwF
& 71.7 & 67.0 & 65.1 & 59.9
& 38.3 & 34.8 & 34.7 & 32.3
& 47.0 & 45.9 & 45.8 & 44.2
& -- & -- & -- \\
UCL-GV
& 71.8 & 66.4 & 70.7 & 68.6
& 38.8 & 34.8 & 34.3 & 31.7
& 42.7 & 41.5 & 42.5 & 42.0
& 100 & 100 & 100 \\
ConDA
& 71.6 & 66.4 & 70.6 & 68.5
& 38.1 & 34.3 & 33.9 & 31.1
& 45.5 & 44.4 & 44.9 & 43.6
& 100 & 100 & 100 \\
CoTTA
& 72.2 & 67.6 & 69.2 & 64.7
& 39.3 & 35.5 & 34.7 & 32.1
& 47.4 & 46.3 & 46.1 & 44.6
& -- & -- & -- \\
ReSNT
& 70.7 & 66.2 & 71.3 & 69.4
& 37.2 & 33.3 & 33.8 & 31.1
& 45.5 & 44.5 & 45.5 & 44.7
& 100 & 100 & 100 \\
BrainUICL
& 74.9 & 69.9 & 74.0 & 72.0
& 40.3 & 36.8 & 36.0 & 33.9
& 48.4 & 47.5 & 48.7 & 48.3
& 100 & 100 & 100 \\
ADaCoRe
& \textbf{77.6} & \textbf{71.6} & \textbf{75.5} & \textbf{73.3}
& \textbf{55.6} & \textbf{54.0} & \textbf{51.4} & \textbf{51.4}
& \textbf{61.7} & \textbf{58.2} & \textbf{72.2} & \textbf{70.6}
& $\sim 15$ & $\sim 10$ & $\sim 5$ \\
\bottomrule
\end{tabular}}
\end{table*}

\subsection{Exemplar Maintenance via Prototype-Confidence Selection}
\label{subsec:pcs}

To maintain a fixed memory budget, we manage the replay buffer using class- and confidence-aware selection strategies for true and pseudo-labeled samples. Let $\varphi(x)$ denote the \emph{frozen} feature extractor to represent each stored EEG segment.
For each class $y$, we compute its feature-space prototype:
\[
\boldsymbol{\mu}_y = \frac{1}{|\mathcal{S}_y|} \sum_{x \in \mathcal{S}_y} \varphi(x),
\]
where $\mathcal{S}_y$ is the set of stored samples for class $y$. If the true-labeled partition exceeds its quota $M_{\text{true}}$, we rank samples by $\|\varphi(x)-\boldsymbol{\mu}_y\|_2$ and retain those closest to $\boldsymbol{\mu}_y$.

A pseudo-labeled sample is accepted into the buffer only if its maximum softmax confidence exceeds a threshold $\tau$ for at least $N_{\min}$ \emph{time windows} (default $\tau{=}0.9$, $N_{\min}{=}15$). Among the accepted items, the most confident are retained until the pseudo-labeled quota $M_{\text{pseudo}}$ is satisfied.
All budgets are measured as kept timepoints $|y|+|\mathcal{P}|$ (times channels), directly reflecting storage cost. Pseudo-labeled samples are stored in the same compressed form with metadata $(u,d,\mathcal{P})$. The full ADaCoRe algorithm is provided in Appendix.

\section{Experiment}

In this section, we first describe the experimental setups and then present the experimental results with an in-depth analysis.

\subsection{Experimental Setups}

\textbf{Datasets.}
We evaluate ADaCoRe on three representative EEG benchmarks with diverse temporal and cognitive characteristics: 
\textbf{ISRUC}~\cite{khalighi2016isruc} (overnight sleep staging), \textbf{FACED}~\cite{faced2023} (emotion decoding from naturalistic interactions), and
\textbf{PhysioNet-MI}~\cite{schalk2004bci2000} (motor imagery trials with $\mu$/$\beta$ ERD/ERS). 
Data preprocessing follows the protocol used in previous work~\cite{zhou2025brainuicl}, with dataset-specific filter bands and saliency presets detailed in Appendix.

\textbf{Baselines.}
We benchmark against three groups of baselines: 
(1) \emph{CL/CDA methods} including MMD~\cite{Gretton2006_kernel_method_two_sample_problem}, TSTCC~\cite{eldele2021tstcc}, EWC~\cite{kirkpatrick2017ewc}, LwF~\cite{li2017lwf}, UCL-GV~\cite{Taufique2022_unsupervised_continual_learning}, ConDA~\cite{Taufique2021_conda_unsupervised_domain_adaptation}, CoTTA~\cite{wang2023cotta}, and ReSNT~\cite{Duan2023_replay_stochastic_neural_transformation}; 
(2) \emph{Replay-based prior art}, represented by BrainUICL~\cite{zhou2025brainuicl}; and 
(3) \emph{Internal ablations and variants}, including full signals without compression, polyphase only without saliency protection, and ADaCoRe with both polyphase and keyframe. 
We compare random selection versus prototype-confidence selection (PCS) under matched memory budgets measured in stored timepoints, and report a no-limit upper bound on FACED (135 full signals).

\textbf{Implementation Details.}
All models are initialized from a common source-pretrained encoder and adapted incrementally in a subject-by-subject stream. Unless otherwise noted, we use a 1D-ResNet-18 encoder, followed by a 2-layer GRU and a linear classifier. 
The CPC teacher shares the same backbone and is trained with contrastive predictive coding on source subjects~\cite{oord2018cpc}. 
Saliency stride is $s = 5$, the smoothing window is $W=\lfloor 0.5\,F_s\rfloor$ in samples, and the Kaiser parameter is $\beta=8.6$~\cite{kaiser1974window}. 
Protection windows span 0.5-1\,s around salient peaks, with coverage cap $\phi = 0.05$, and median/MAD threshold $\kappa = 2.5$~\cite{rousseeuw1993mad}. 
Pseudo-labeled samples are accepted only if their confidence exceeds $\tau = 0.9$ for at least $N_{\min} = 15$ time windows. See Appendix for further details.

\textbf{Evaluation Metrics.}
Following previous work~\cite{zhou2025brainuicl}, we report \emph{Accuracy} (\ACC) and \emph{Macro-averaged F1} (\MFone) as \textit{plasticity} metrics, and \emph{Average Accuracy on future (not-yet-adapted) subjects} (\AAA) and \emph{Average Macro-F1 on future subjects} (\AAFone) as \textit{stability} metrics. 
Formal definitions are provided in Appendix.

\subsection{Experimental Results}

\textbf{Overall Performance.}
Table~\ref{tab:overall_perf} reports UICL results on ISRUC, FACED, and PhysioNet--MI. ADaCoRe outperforms both generic CL/CDA baselines and the state-of-the-art BrainUICL~\cite{zhou2025brainuicl} in all metrics, showing consistent gains in both plasticity and stability across all three datasets. On ISRUC, the improvements over BrainUICL reach $+2.7$\% ACC and $+1.3$\% AAF1. On FACED, which involves more variable affective contents, improvements reach $+15.3$\% ACC and $+17.5$\% AAF1, suggesting that our signal-aware compression, reconstruction, and maintenance strategy better captures meaningful variation under limited memory. On PhysioNet--MI, a motor-imagery decoding task with shorter trials, ADaCoRe still yields substantial gains (up to $+13.3$\% ACC and $+22.3$\% AAF1), indicating that the proposed framework generalizes beyond sleep staging and affective EEG to different paradigms while operating with only 5-15\% of the BrainUICL buffer.

\textbf{Ablation Study.}
To isolate the contribution of each component, we evaluate four compression variants and three buffer-selection strategies in Table~\ref{tab:ablation_study_compression}.
For \textbf{compression and reconstruction}, we compare 
(1) full signal storage without compression; 
(2) \emph{linear-only} compression without saliency-aware protection; 
(3) polyphase-only compression without saliency-aware protection; and 
(4) ADaCoRe with both polyphase compression and keyframe preservation. 
For \textbf{buffer management} under a fixed memory budget (buffer size equivalent to 135 full samples), we further compare 
(5) an oracle-like ``no buffer limit'' baseline that stores all past data; 
(6) random selection; and 
(7) prototype-confidence selection (PCS).

On FACED, a naive linear compression baseline severely degrades both plasticity and stability (\eg, ACC drops to $22.6$\% and AAF1 to $14.6$\%), indicating that simply downsampling in the time domain destroys task-relevant EEG structure. In contrast, the frequency-aware polyphase design maintains performance close to the full-storage setting (ACC $52.9$\% vs.\ $54.3$\%), underscoring the importance of spectrally structured compression. Adding saliency-aware keyframe preservation on top of polyphase (ADaCoRe) further improves both plasticity and stability, achieving the best overall performance under the same number of stored points. Under a fixed buffer size, PCS dramatically outperforms random selection and even surpasses the no-buffer-limit baseline, highlighting that selecting samples based on both prototype distance and confidence is crucial under tight memory constraints.

\begin{figure}[!t]
  \centering
  \includegraphics[width=0.98\linewidth]{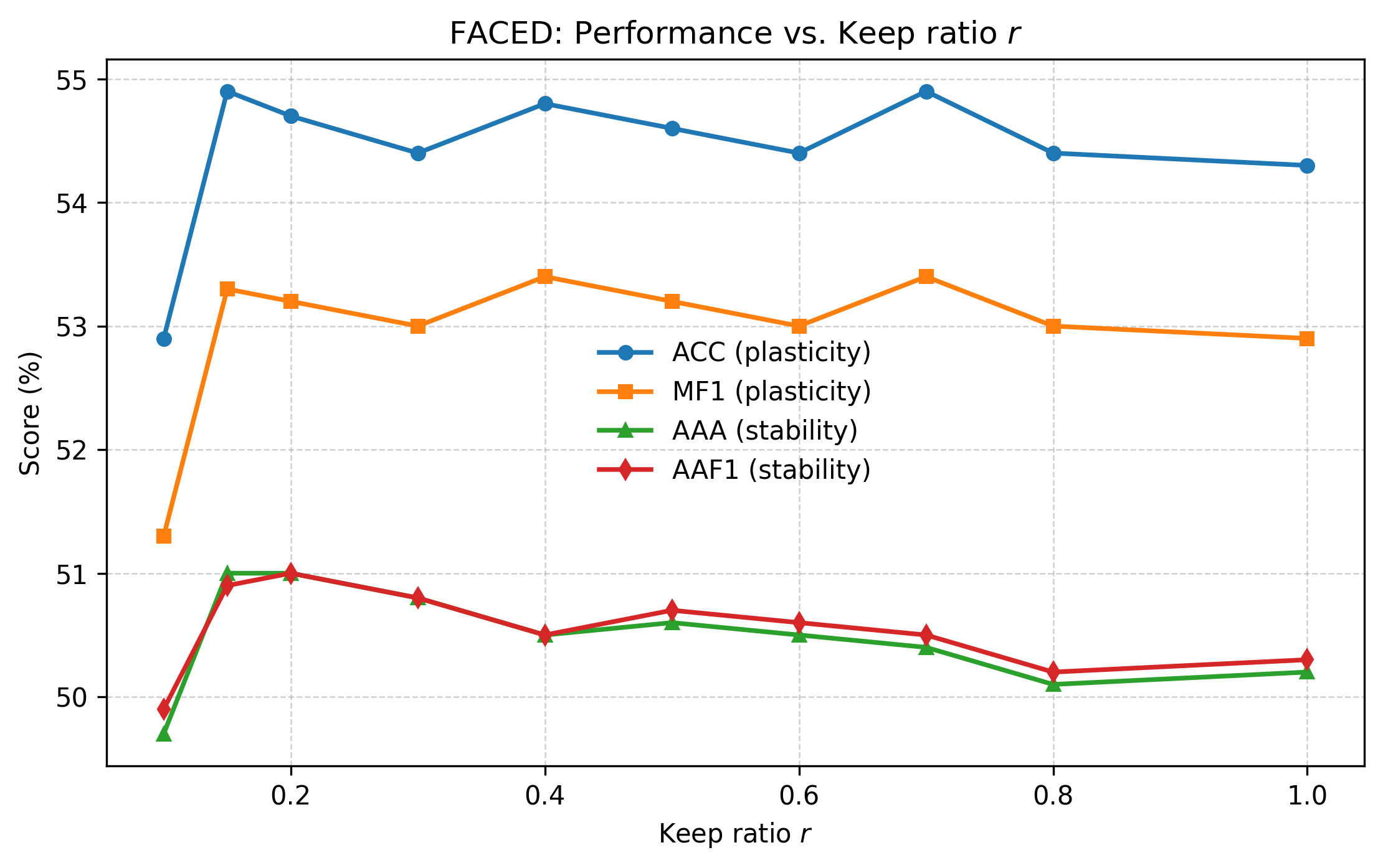}
  \caption{Effect of the keep ratio \(r\) under polyphase-only compression on \textbf{FACED}.}
  \label{fig:ratio_visuals}
\end{figure}

\begin{figure*}[!t]
  \centering
  \subfloat[ISRUC reconstruction grid.\label{fig:recon_isruc}]{%
    \includegraphics[width=0.32\textwidth]{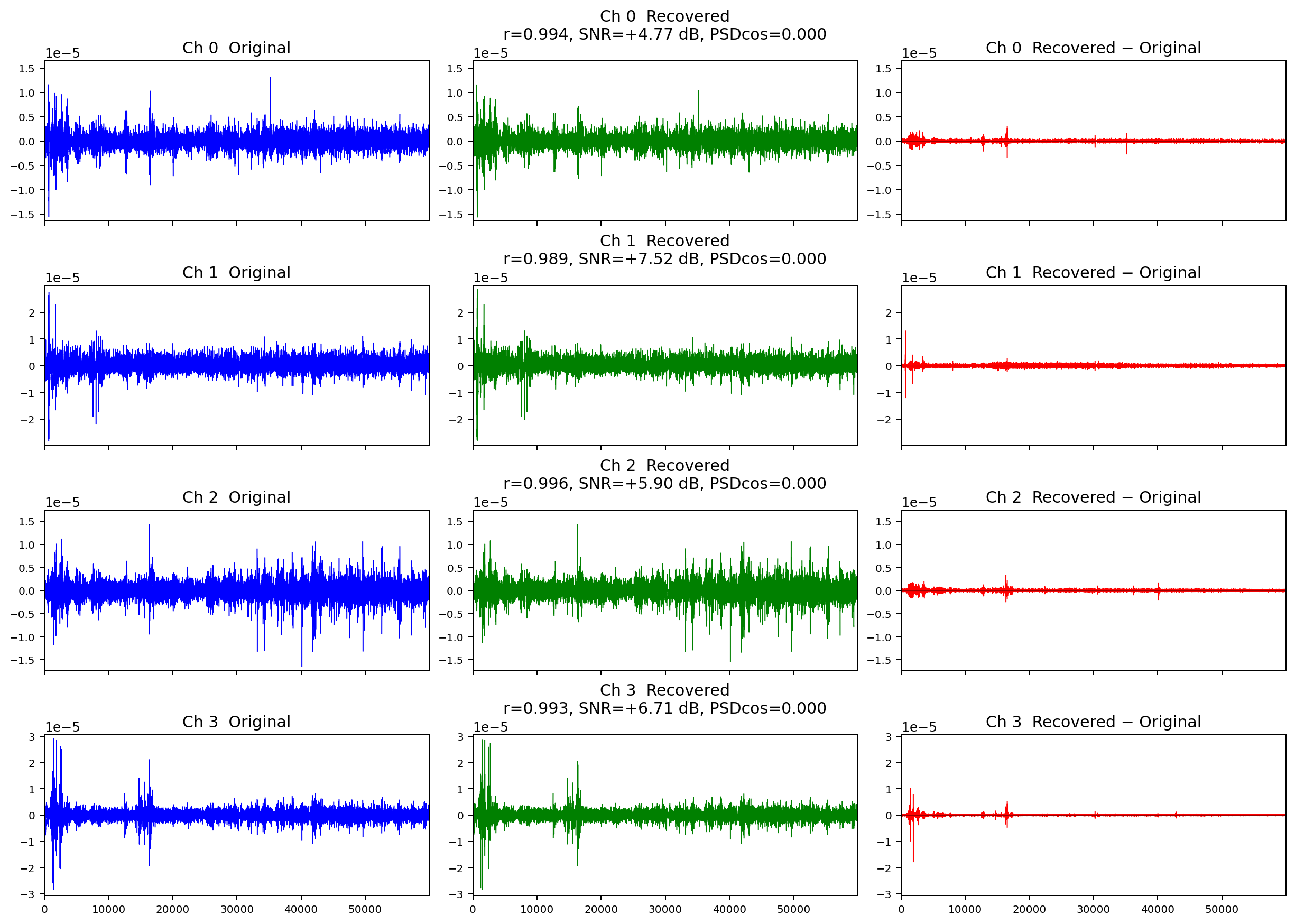}}
  \hfill
  \subfloat[FACED reconstruction grid.\label{fig:recon_faced}]{%
    \includegraphics[width=0.32\textwidth]{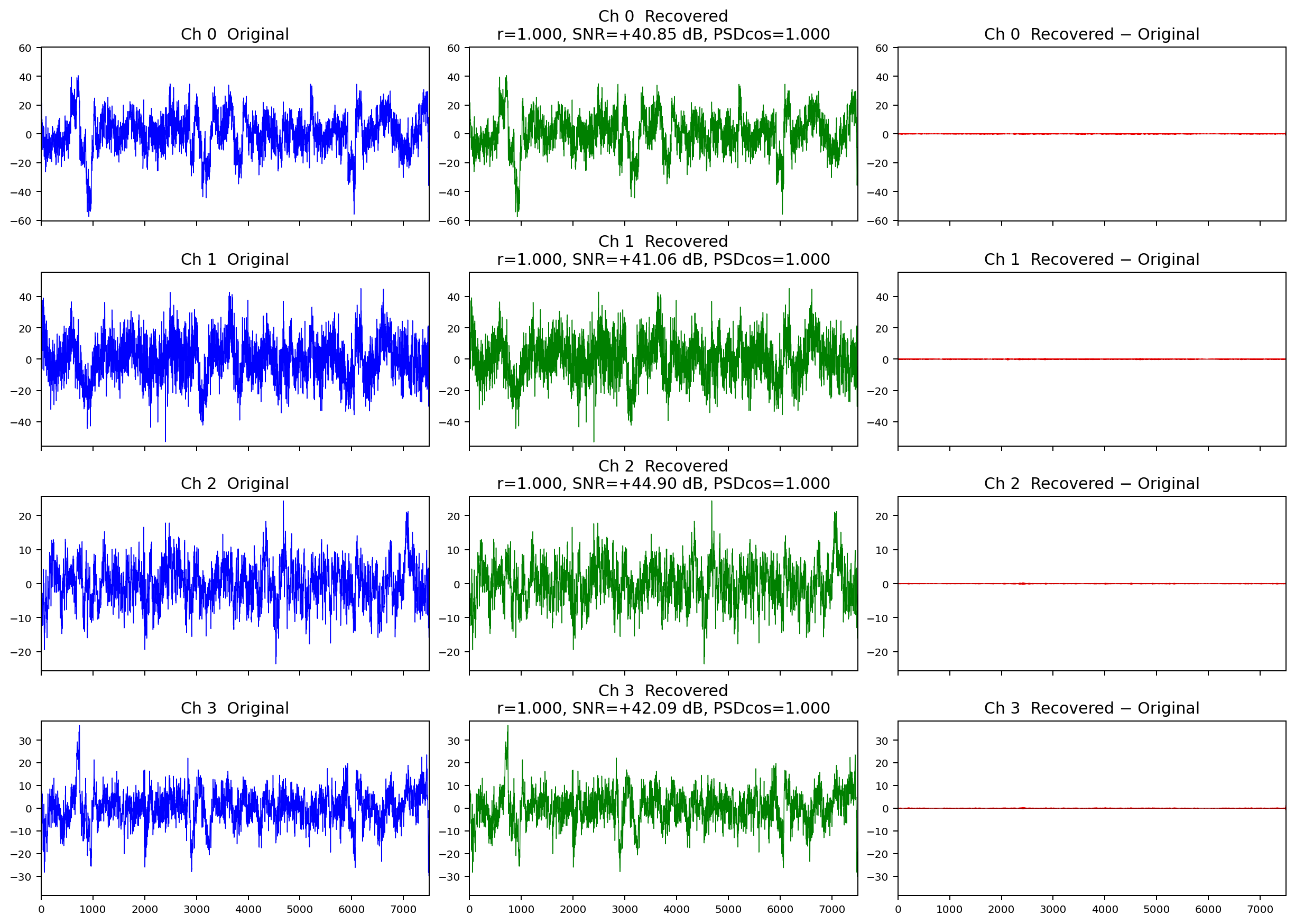}}
  \hfill
  \subfloat[PhysioNet--MI reconstruction grid.\label{fig:recon_mi}]{%
    \includegraphics[width=0.32\textwidth]{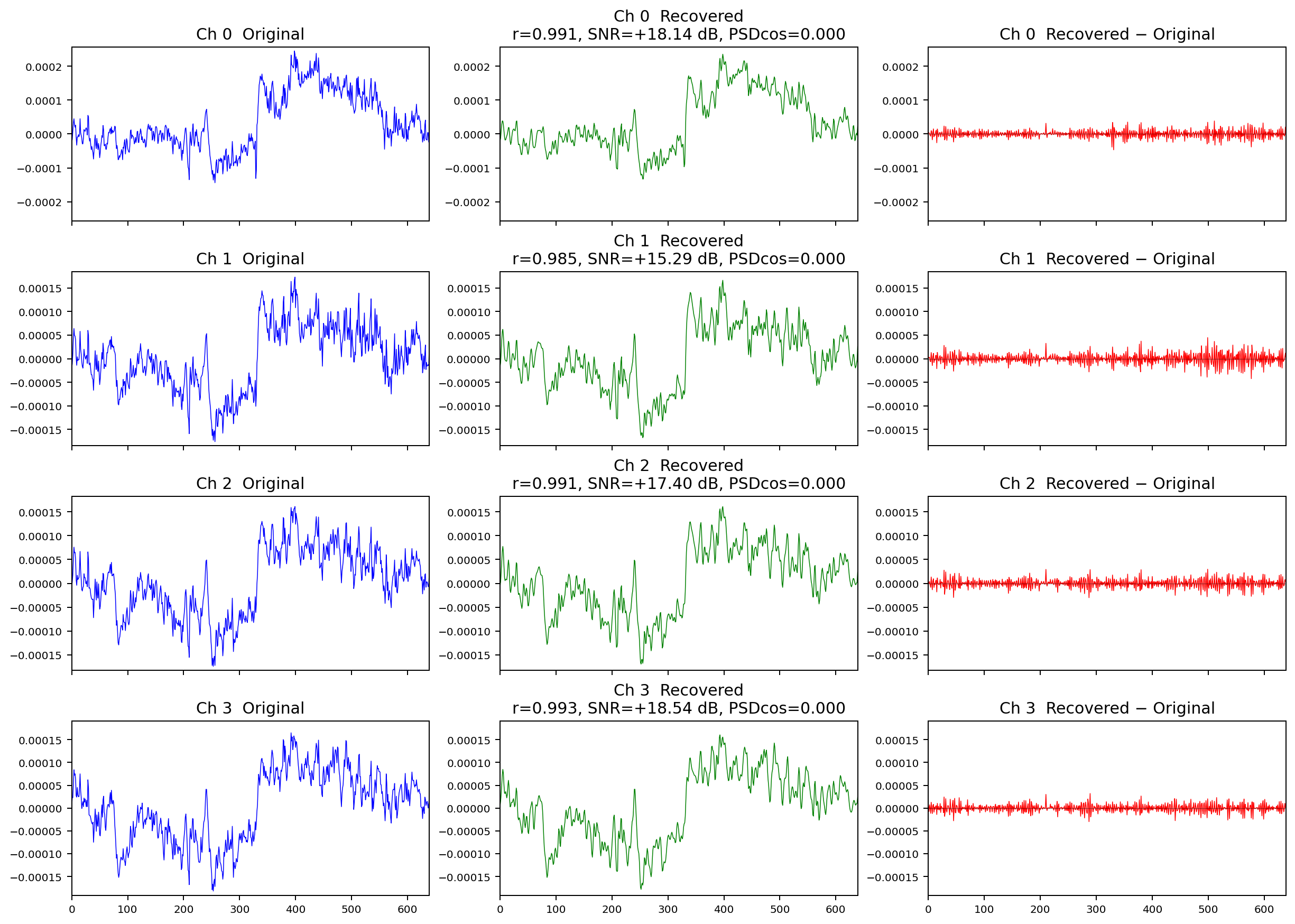}}
  \caption{Qualitative reconstruction fidelity with ADaCoRe. For each dataset, we show a representative segment from the first few EEG channels (rows) and three traces per channel (columns): original waveform, waveform reconstructed from the compressed representation, and pointwise reconstruction error (Recovered $-$ Original). The title of each original panel reports three agreement measures between the original signal $x$ and its reconstruction $\tilde{x}$: $r$ (Pearson correlation over time), SNR in dB (signal power vs.\ reconstruction error power), and PSDcos (cosine similarity between Welch power spectra). Higher $r$, SNR, and PSDcos indicate better preservation of waveform shape, amplitude, and frequency content.}
  \label{fig:recon_visuals}
\end{figure*}

\textbf{Visualization Analysis.}
Figure~\ref{fig:ratio_visuals} plots the four evaluation metrics on FACED as a function of keep ratio $r$ under polyphase-only compression. Performance is stable for $r\gtrsim 0.15$, with a broad plateau where plasticity and stability are near-optimal; extremely aggressive compression ($r=0.1$) slightly harms performance, while moving towards $r=1$ provides little benefit and can reduce stability, suggesting a regularizing effect of moderate compression. Figure~\ref{fig:recon_visuals} provides qualitative examples of reconstruction fidelity on all three datasets: reconstructed traces closely follow originals, and error panels mainly contain low-amplitude residuals, with high per-channel correlation, SNR, and spectral similarity.

\begin{table}[!t]
\caption{Ablation of compression, reconstruction, and prototype-confidence selection (PCS) on \textbf{FACED}. The buffer size for PCS is set to 135 full samples.}
\label{tab:ablation_study_compression}
\label{tab:ablation_study_selection}
\centering
\resizebox{0.95\linewidth}{!}{%
\begin{tabular}{lcccc}
\toprule
\multirow{2}{*}{Methods}
& \multicolumn{2}{c}{\textit{Plasticity}}
& \multicolumn{2}{c}{\textit{Stability}} \\
\cmidrule(lr){2-3}\cmidrule(lr){4-5}
& \ACC & \MFone & \AAA & \AAFone \\
\midrule
\multicolumn{5}{l}{\emph{Compression and reconstruction}} \\
Without compression
& 54.3 & 52.9 & 50.2 & 50.3 \\
Linear only (no keyframe)
& 22.6 & 16.1 & 20.4 & 14.6 \\
Polyphase only (no keyframe)
& 52.9 & 51.3 & 49.7 & 49.9 \\
ADaCoRe (polyphase + keyframe)
& \textbf{55.6} & \textbf{54.0} & \textbf{51.4} & \textbf{51.4} \\
\midrule
\multicolumn{5}{l}{\emph{Prototype-confidence selection (PCS)}} \\
No buffer limit
& 40.3 & 36.8 & 36.0 & 33.9 \\
Random selection
& 16.8 & 9.7 & 17.9 & 11.3 \\
PCS (prototype + confidence)
& \textbf{54.5} & \textbf{53.1} & \textbf{50.3} & \textbf{50.4} \\
\bottomrule
\end{tabular}}
\end{table}

\section{Conclusion}
In this work, we present \textbf{ADaCoRe}, a signal-aware unsupervised continual learning approach for EEG that jointly integrates saliency-driven keyframe protection, rational polyphase compression, overwrite-based reconstruction, and prototype--confidence selection under strict memory constraints. Through principled design and analysis, ADaCoRe achieves strong plasticity--stability trade-offs while significantly reducing replay storage. Experiments across sleep staging, motor imagery, and affective decoding demonstrate consistent advantages over recent strong baselines. Future directions include scaling to richer affective datasets, adaptive memory allocation under nonstationary shifts, and extending the framework to other biosignals and on-device applications.

\clearpage

\nocite{crochiere1983multirate,oppenheim2010dsp,maragos1993amfm,maragos1993energy,rousseeuw1993mad}
\bibliographystyle{IEEEtran}
\bibliography{original/icme2025references}

\clearpage
\appendices

\section{Signal Processing Details}
\label{app:sigproc}

This supplement details the discrete-time model and signal-processing operators that underlie our saliency, compression, and reconstruction modules.

\subsection{Discrete-time model and notation}
Let $F_s$ denote the sampling rate in hertz. A continuous-time signal $x_a(t)$ is observed as a discrete-time sequence
\begin{equation}
x[n] \;=\; x_a\!\left(\frac{n}{F_s}\right), \qquad n\in\mathbb{Z}.
\end{equation}
For $C$ channels, we write $\mathbf{x}[n]\in\mathbb{R}^{C}$. In typical implementations, a tensor of shape $[T,C,L]$ corresponds to $T$ epochs, $C$ channels, and $L$ samples per epoch.

\subsection{Linear time-invariant filtering}
A linear time-invariant (LTI) filter with impulse response $h[n]$ acts on a discrete-time signal $x[n]$ via convolution:
\begin{equation}
y[n] \;=\; (h * x)[n] \;=\; \sum_{k=-\infty}^{\infty} h[k]\;x[n-k].
\end{equation}
In our implementation we employ finite-impulse-response (FIR) and infinite-impulse-response (IIR) designs with passbands chosen according to the downstream task. Zero-phase responses are implemented by forward--backward filtering (\texttt{filtfilt}), which applies $h$ in both directions to cancel phase distortion~\cite{oppenheim2010dsp}.

\subsection{Sampling, resampling, and polyphase realization}
If the underlying continuous-time signal $x_a$ is effectively band-limited to $B$\,Hz, sampling above $2B$ avoids aliasing. Any rate reduction by an integer factor $M$ lowers the Nyquist to $F_s/(2M)$; consequently, an anti-alias low-pass filter is required before decimation.

We define the upsampler $(\uparrow L)$ and decimator $(\downarrow M)$ acting on $x[n]$ by
\begin{equation}
\begin{aligned}
(\uparrow L\, x)[n] &= 
\begin{cases}
x[n/L], & n \equiv 0 \ (\mathrm{mod}\ L),\\
0, & \text{otherwise},
\end{cases} \\
(\downarrow M\, x)[n] &= x[nM].
\end{aligned}
\end{equation}
Let $h_{L,M}[n]$ be a real, linear-phase, low-pass FIR kernel that serves simultaneously as an anti-imaging and anti-aliasing filter. The rational polyphase resampler is
\begin{equation}
\label{eq:frp}
f_{\mathrm{RP}}(x;L,M) \;=\; \downarrow M \,\big( \, h_{L,M} * (\uparrow L\, x) \,\big).
\end{equation}
Efficient polyphase and multirate implementations follow standard references~\cite{crochiere1983multirate,oppenheim2010dsp}.

\subsection{Bandpass filtering and analytic envelopes}
Task-relevant EEG rhythms (\eg, delta, theta, alpha ($\mu$), beta, sigma) are isolated by zero-phase bandpass filters. For a band-limited signal $x[n]$ in a given band, the analytic signal via the Hilbert transform $\mathcal{H}\{\cdot\}$ is
\begin{equation}
z[n] \;=\; x[n] + \mathrm{j}\,\mathcal{H}\{x[n]\},
\end{equation}
and the envelope $e[n]=|z[n]|$ serves as a smooth estimate of instantaneous amplitude; $e[n]^2$ approximates instantaneous band power~\cite{oppenheim2010dsp}. For multichannel data, envelopes are smoothed and averaged across channels before they are combined into the saliency trace.

\subsection{Transient energy via the Teager--Kaiser operator}
Transient events are emphasized by the Teager--Kaiser energy operator
\begin{equation}
\Psi\!\big(x[n]\big) \;=\; x[n]^2 \;-\; x[n-1]\,x[n+1],
\end{equation}
which is applied per channel, rectified, short-window smoothed, and aggregated across channels~\cite{maragos1993amfm,maragos1993energy}. The resulting transient energy sequence is combined with band-power envelopes to form the scalar saliency trace $S[n]$ described in the main text.

\section{Metric Definitions}
\label{app:metrics}

Let the unlabeled subject stream be $\{\mathcal{U}^{(t)}\}_{t=1}^{T}$ and $M_t$ denote the model after adapting on subject $t$.

\subsection{Plasticity (current subject)}
We quantify plasticity by averaging performance on the current subject at each adaptation step:
\begin{align}
\ACC \;&=\; \frac{1}{T}\sum_{t=1}^{T}\mathrm{Acc}\big(M_t,\ \mathcal{U}^{(t)}\big),\\
\MFone \;&=\; \frac{1}{T}\sum_{t=1}^{T}\mathrm{MacroF1}\big(M_t,\ \mathcal{U}^{(t)}\big).
\end{align}

\subsection{Stability and generalization (future subjects)}
For stability/generalization, at each step $t$ we evaluate $M_t$ on subjects that have not yet been adapted, collected into $G_t=\{s\mid s>t\}$. We then average per-step performance over $G_t$ and over $t$:
\begin{align}
\AAA \;&=\; \frac{1}{T-1}\sum_{t=1}^{T-1}\ \frac{1}{|G_t|}\sum_{s\in G_t}\mathrm{Acc}\big(M_t,\ \mathcal{U}^{(s)}\big),\\
\AAFone \;&=\; \frac{1}{T-1}\sum_{t=1}^{T-1}\ \frac{1}{|G_t|}\sum_{s\in G_t}\mathrm{MacroF1}\big(M_t,\ \mathcal{U}^{(s)}\big).
\end{align}
When a dataset defines official evaluation folds, we follow that protocol and compute all metrics within each fold.

\section{Dataset-specific defaults and differences}
\label{app:ds-defaults}

\begin{table*}[!t]
\caption{Dataset-specific defaults for saliency, protection, and preprocessing.}
\label{tab:ds_defaults}
\centering
\resizebox{\linewidth}{!}{%
\begin{tabular}{lccccccccc}
\toprule
Dataset & $F_s$ & Bands (examples) & Top-$K$ & Robust stat. & $\gamma$ & $\phi$ & $\rho$ (s)  & $W$ (s) \\
\midrule
ISRUC & 100 & $\{[0.5,4], [11,16]\}$ & 2 & median & 0.3 & 0.05 & 0.5--1.0 & 0.5 \\
FACED & 250 & $\{[0.5,4], [4,8], [8,13], [13,30], [30,45]\}$ & 3 & median & 0.1 & 0.10 & 0.5--1.0 & 0.5 \\
PhysioNet--MI & 160 & $\{[6.0, 9.0], [8.0, 13.0], [13.0, 30.0]\}$ & 2 & trimmed mean (0.1) & 0.3 & 0.10 & 0.1--0.5 & 0.1 \\
\bottomrule
\end{tabular}}
\\[0.3em]
\footnotesize Peak detection uses $\kappa{=}2.5$ unless stated; ISRUC: skip non-EEG channels (\eg, EOG=$2$) when present.
\end{table*}

\section{Additional Experimental Analyses}
\label{app:add-exp}

\subsection{Compute and latency}
\label{app:latency}

Table~\ref{tab:app_latency_compute} reports the model compute and CPU-side replay latency on FACED for a 30\,s segment with shape $32\times7500$ and batch size $B{=}1$. The encoder contains 11.506M parameters and requires $8.689\times10^{8}$ MACs per forward pass, corresponding to approximately 1.738 GFLOPs if one MAC is counted as two floating-point operations. At replay time, ADaCoRe adds reconstruction before the standard forward pass. This increases the per-segment latency from 17.554\,ms to 20.480\,ms with one CPU thread, and from 7.903\,ms to 10.277\,ms with 24 CPU threads. The resulting additional replay-time overhead is 2.926/2.374\,ms per 30\,s segment, while the one-time compression cost incurred when inserting a segment into the buffer is approximately 88--90\,ms.

\begin{table}[!t]
\caption{Compute and CPU latency summary on FACED for a 30\,s segment with shape $32\times7500$ and $B{=}1$.}
\label{tab:app_latency_compute}
\centering
\footnotesize
\setlength{\tabcolsep}{4.2pt}
\resizebox{\columnwidth}{!}{%
\begin{tabular}{lc}
\toprule
Item & Value \\
\midrule
\#Params (THOP) & 11.506 M \\
MACs / forward (THOP) & $8.689\times 10^{8}$ \\
Approx.\ GFLOPs / forward ($2\times$MACs) & 1.738 \\
\midrule
BrainUICL replay-time forward (raw, 1 thread / 24 threads) & 17.554 ms / 7.903 ms \\
ADaCoRe replay E2E (recover+forward, 1 thread / 24 threads) & 20.480 ms / 10.277 ms \\
Replay overhead (ADaCoRe$-$BrainUICL) & 2.926 ms / 2.374 ms \\
Compression insertion cost (one-time, per segment) & $\sim$88--90 ms \\
\bottomrule
\end{tabular}}
\end{table}

\subsection{Sensitivity to keyframe hyperparameters}
\label{app:keyframe-sensitivity}

\begin{figure*}[!t]
\centering
\subfloat[Sweep of the coverage cap $\phi$ with $\rho{=}0.5$\,s.\label{fig:app_phi_sweep}]{%
  \includegraphics[width=0.48\textwidth]{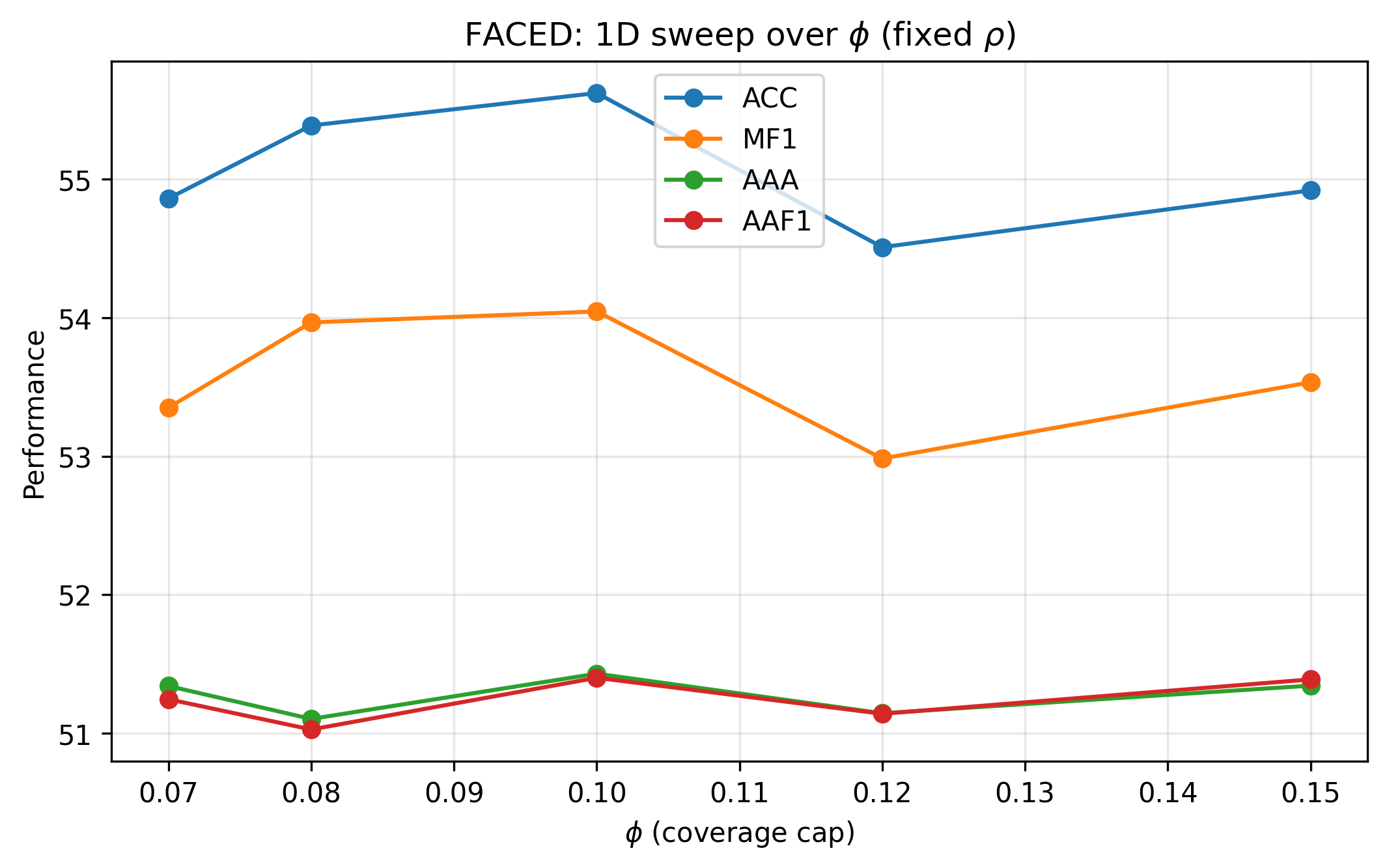}}
\hfill
\subfloat[Sweep of the keyframe radius $\rho$ with $\phi{=}0.1$.\label{fig:app_rho_sweep}]{%
  \includegraphics[width=0.48\textwidth]{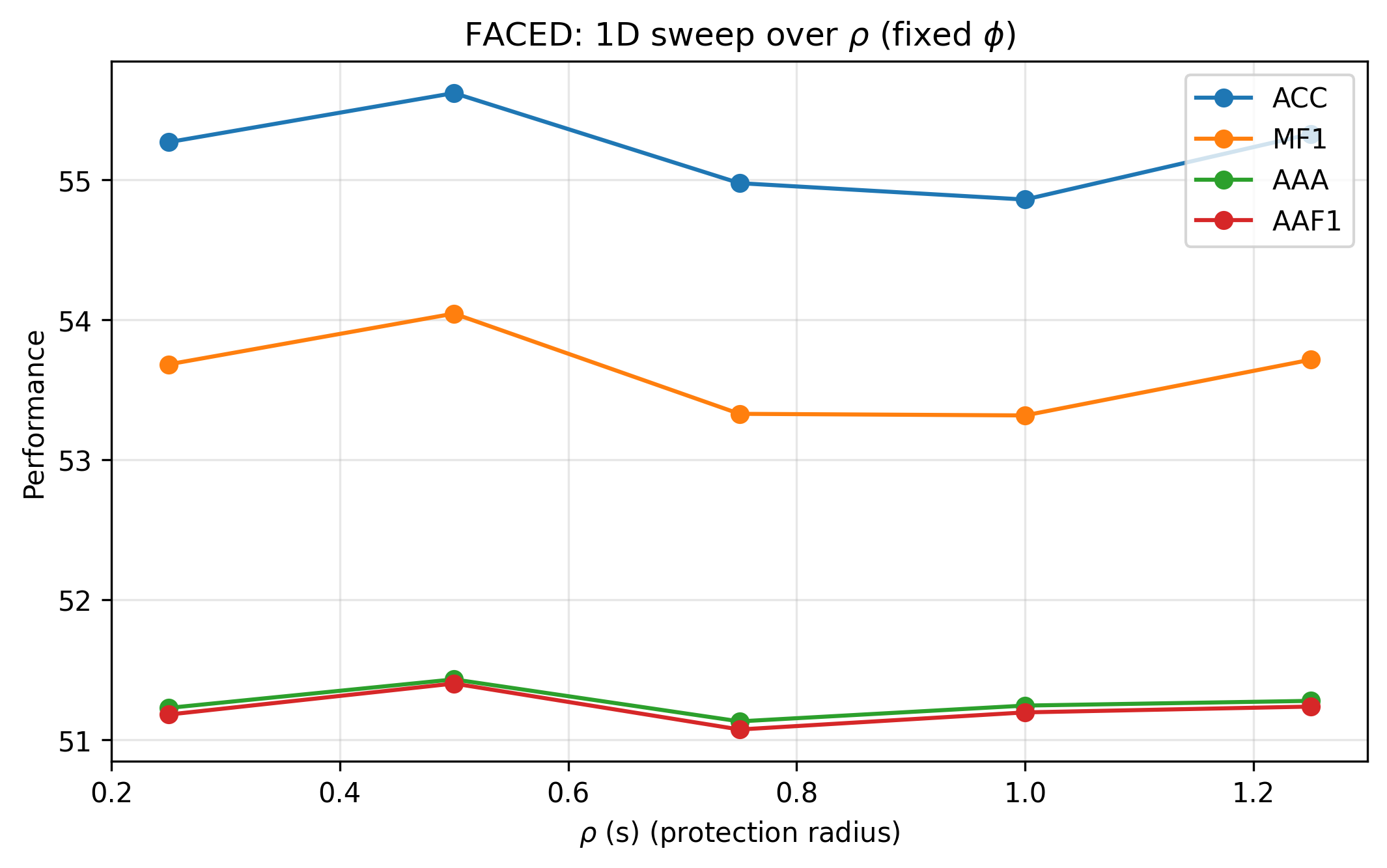}}
\caption{FACED sensitivity for keyframe hyperparameters. Left: $\phi\in[0.07,0.15]$ with $\rho{=}0.5$\,s. Right: $\rho\in[0.25,1.25]$\,s with $\phi{=}0.1$.}
\label{fig:app_phi_rho_sweep}
\end{figure*}

Figure~\ref{fig:app_phi_rho_sweep} evaluates the sensitivity of the keyframe coverage cap $\phi$ and neighborhood radius $\rho$ on FACED. When sweeping $\phi\in[0.07,0.15]$ with $\rho{=}0.5$\,s, ACC/MF1 varies within approximately one point and AAA/AAF1 within less than 0.4 points. When sweeping $\rho\in[0.25,1.25]$\,s with $\phi{=}0.1$, ACC/MF1 varies by at most about 0.8 points and AAA/AAF1 by at most about 0.35 points. These results indicate a broad stable plateau around the default setting $(\phi{=}0.1,\rho{=}0.5\,\mathrm{s})$.

\subsection{Robustness to EMG-like artifacts}
\label{app:noise-robustness}

Table~\ref{tab:app_noise_robustness} evaluates whether Teager--Kaiser-based saliency is vulnerable to EMG-like high-frequency artifacts on FACED. We use \texttt{bandpass\_gaussian} noise with seed 4321 under two modes. In \texttt{compress\_only}, noise is injected only during buffer insertion, \ie, during saliency estimation and compression, isolating the robustness of keyframe selection. In \texttt{all}, noise is applied end-to-end to input, replay, training, and evaluation, reflecting a stronger deployment-time corruption. Under \texttt{compress\_only}, degradation remains negligible from 40\,dB to 5\,dB, with a maximum ACC/MF1 drop no larger than 0.7 and nearly unchanged AAA/AAF1. Under \texttt{all}, performance is stable at moderate noise levels from 40\,dB to 10\,dB, while the extreme 5\,dB setting causes a limited drop of 1.7 points in ACC/MF1. This suggests that band-limited processing and morphology-aware keyframe protection prevent high-frequency artifacts from dominating replay selection, and the remaining degradation under severe noise mainly reflects general input corruption rather than a saliency-specific failure mode.

\begin{table}[!t]
\caption{Noise robustness under EMG-like artifacts on FACED using \texttt{bandpass\_gaussian} noise with seed 4321.}
\label{tab:app_noise_robustness}
\centering
\footnotesize
\setlength{\tabcolsep}{3.2pt}
\resizebox{\columnwidth}{!}{%
\begin{tabular}{llcccc}
\toprule
\multirow{2}{*}{Noise} & \multicolumn{2}{c}{\texttt{compress\_only}} & \multicolumn{2}{c}{\texttt{all}} & \multirow{2}{*}{$\Delta$ACC/MF1} \\
\cmidrule(lr){2-3}\cmidrule(lr){4-5}
& ACC/MF1 & AAA/AAF1 & ACC/MF1 & AAA/AAF1 & \\
\midrule
Clean  & 55.6/54.0 & 51.4/51.4 & 55.6/54.0 & 51.4/51.4 & 0.0/0.0 \\
+40 dB & 55.0/53.4 & 51.4/51.3 & 55.2/53.7 & 51.4/51.4 & $-0.4/-0.3$ \\
+20 dB & 54.9/53.3 & 51.3/51.3 & 55.6/54.1 & 51.4/51.4 & $+0.0/+0.1$ \\
+10 dB & 55.0/53.4 & 51.4/51.3 & 55.3/53.7 & 51.0/51.0 & $-0.3/-0.3$ \\
+5 dB  & 55.4/54.0 & 51.1/51.1 & 53.9/52.3 & 50.2/50.1 & $-1.7/-1.7$ \\
\bottomrule
\end{tabular}}
\end{table}

\begin{algorithm}[t]
\caption{ADaCoRe (per-sequence processing and maintenance)}
\label{alg:adacore}
\begin{algorithmic}
\REQUIRE EEG segment $x\in\mathbb{R}^{N\times C}$, target keep ratio $r$, budgets $M_{\text{true}}, M_{\text{pseudo}}$ (kept points)

\STATE \textbf{A. Saliency computation}
\STATE Compute scalar saliency trace $S[n]$ from band-limited envelopes and Teager--Kaiser energy.
\STATE Detect peaks above a median- and MAD-based threshold, expand each by radius $\rho$, and form protected set $\mathcal{P}$ (cap $|\mathcal{P}|\le\phi N$ and include endpoints).

\STATE \textbf{B. Compression}
\STATE Choose rational pair $(u,d)$ with $d\le d_{\max}$ such that $u/d\simeq r$ (optionally refined via nearby Farey neighbors).
\STATE Compress $x$ by rational polyphase resampling $y = f_{\mathrm{RP}}(x;u,d)$ and store $(y,\{x[t]\}_{t\in\mathcal{P}},u,d,\mathcal{P})$ in the replay buffer.

\STATE \textbf{C. Reconstruction at replay time}
\STATE Reconstruct dense sequence $\tilde{x} = f_{\mathrm{RP}}(y;d,u)$ and overwrite $\tilde{x}[t]\leftarrow x[t]$ for all $t\in\mathcal{P}$.

\STATE \textbf{D. Exemplar maintenance (PCS)}
\IF{pseudo-labeled exemplar passes the confidence gate on $\ge N_{\min}$ time windows}
    \STATE Insert its compressed tuple into the pseudo-labeled partition of the buffer.
\ENDIF
\IF{true-labeled partition exceeds budget $M_{\text{true}}$}
    \STATE Update per-class prototypes $\boldsymbol{\mu}_y$ and retain exemplars closest to $\boldsymbol{\mu}_y$ until the budget is met.
\ENDIF
\IF{pseudo-labeled partition exceeds budget $M_{\text{pseudo}}$}
    \STATE Rank pseudo-labeled exemplars by mean confidence and keep the top ones under the timepoint budget.
\ENDIF

\end{algorithmic}
\end{algorithm}

\section{\texorpdfstring{Robust adaptive thresholding: why Med + $k\times 1.4826$ MAD?}{Robust adaptive thresholding: why Med + k x 1.4826 MAD?}}
\label{app:mad-thresh}

We adopt a robust threshold $\tau=\mathrm{Med}(S)+k\cdot 1.4826\,\mathrm{MAD}(S)$ for peak detection in the saliency trace $S[n]$, where
\[
\mathrm{MAD}(S) \;=\; \operatorname{median}\{|S[n]-\mathrm{Med}(S)|\}.
\]
This construction mirrors the familiar ``$\mu+k\sigma$'' rule while replacing the mean and standard deviation with their robust counterparts (median and a robust scale estimator), yielding high resistance to outliers and heavy-tailed noise---common in EEG due to artifacts such as EMG bursts.

\subsection{Origin of 1.4826 (normal-consistent scaling)}
For a standard normal variable $Z\sim\mathcal{N}(0,1)$,
\[
\operatorname{median}(|Z|)=m \quad \text{such that} \quad \Pr(|Z|\le m)=0.5.
\]
Hence $2\Phi(m)-1=0.5 \Rightarrow \Phi(m)=0.75 \Rightarrow m=\Phi^{-1}(0.75)\approx 0.67448975$,
where $\Phi$ is the standard normal CDF. For $X\sim\mathcal{N}(\mu,\sigma^2)$,
\begin{align*}
\mathrm{MAD}(X) &= \operatorname{median}|X-\mathrm{Med}(X)| \\
              &= \sigma\cdot \operatorname{median}|Z| \\
              &\approx 0.67448975\,\sigma.
\end{align*}
Therefore a \emph{normal-consistent} estimator of $\sigma$ is
\[
\widehat{\sigma}_{\mathrm{MAD}} \;=\; \frac{1}{0.67448975}\,\mathrm{MAD}\ \approx\ 1.4826\,\mathrm{MAD}.
\]
Using $\widehat{\mu}=\mathrm{Med}$ and $\widehat{\sigma}=\widehat{\sigma}_{\mathrm{MAD}}$ makes
\[
\tau \;=\; \widehat{\mu} + k\,\widehat{\sigma}
\]
a robust analogue of the $k\sigma$ threshold (see also \cite{rousseeuw1993mad}). Typical choices $k\in[2,3]$ correspond, under near-normal $S$, to single-sided exceedance probabilities on the order of $\sim2.3\%$ ($k{=}2$) down to $\sim0.13\%$ ($k{=}3$), providing an interpretable knob for controlling false peaks while preserving robustness to outliers.

\subsection{Why not mean and variance?}
The mean and standard deviation have low breakdown points and are easily inflated by transient artifacts, which would raise the threshold and suppress true EEG events. The median and MAD have 50\% breakdown points, keeping thresholds stable across subjects and sessions with heterogeneous noise.

\section{ADaCoRe Algorithm}
\label{app:algorithm}

Algorithm~\ref{alg:adacore} presents the complete ADaCoRe algorithm.

\section{Engineering notes}
\label{app:eng-notes}

Replay uses mini-batches drawn from the buffer. When both true- and pseudo-labeled items are available, we employ a \emph{fixed} mixing ratio between the two sources for batching (kept constant across methods for fairness). The teacher is set to \texttt{eval} mode to freeze batch-normalization statistics and avoid drift during continual adaptation. Since labels are unavailable during individual adaptation, pseudo-labels and their confidence scores follow the standard BrainUICL-style UICL protocol. ADaCoRe further uses PCS to prioritize pseudo-labeled samples by prediction confidence, reducing error accumulation, and to retain prototype-consistent samples by within-class center distance, improving representativeness and diversity. Together with the explicit coverage cap $\phi$, this design prevents the buffer from being dominated by a narrow set of high-confidence samples and mitigates confirmation-bias effects during extended adaptation. Implementation-level details such as batch sizes, optimizer choices, and learning-rate schedules follow standard practice for 1D convolutional EEG encoders and are held constant across all compared methods.

\end{document}